\documentclass{article}
\usepackage{ijcai23}
\usepackage{bbding}
\usepackage{color}
\usepackage{helvet}
\usepackage{subfigure}
\usepackage{amsfonts}
\usepackage{amsmath}
\usepackage{fancyhdr}
\usepackage{times}
\usepackage{soul}
\usepackage{url}
\usepackage[hidelinks]{hyperref}
\usepackage[utf8]{inputenc}
\usepackage[small]{caption}
\usepackage{graphicx}
\usepackage{amsmath}
\usepackage{amsthm}
\usepackage{booktabs}
\usepackage{algorithm}
\usepackage{algorithmic}
\usepackage{graphicx}
\graphicspath{{figures/}}
\usepackage[switch]{lineno}

\title{Low-Confidence Samples Mining for Semi-supervised Object Detection}

\author{
Guandu Liu$^{1,2}$
\and
Fangyuan Zhang$^{1,2}$\and
Tianxiang Pan$^{1,2}$\and
Jun-Hai Yong$^{1,2}$\And
Bin Wang$^{1,2}$\footnote{Corresponding Author}
\affiliations
$^1$School of Software, Tsinghua University, China\\
$^2$Beijing National Research Center for Information Science and Technology (BNRist), China\\
\emails
\{liugd21, zhangfy19\}@mails.tsinghua.edu.cn,
ptx9363@gmail.com,
\{yongjh, wangbins\}@tsinghua.edu.cn
}

\begin{document}

\maketitle

\begin{abstract}
Reliable pseudo-labels from unlabeled data play a key role in semi-supervised object detection (SSOD). However, the state-of-the-art SSOD methods all rely on pseudo-labels with high confidence, which ignore valuable pseudo-labels with lower confidence. Additionally, the insufficient excavation for unlabeled data results in an excessively low recall rate thus hurting the network training. In this paper, we propose a novel Low-confidence Samples Mining (LSM) method to utilize low-confidence pseudo-labels efficiently. Specifically, we develop an additional pseudo information mining (PIM) branch on account of low-resolution feature maps to extract reliable large-area instances, the IoUs of which are higher than small-area ones. Owing to the complementary predictions between PIM and the main branch, we further design self-distillation (SD) to compensate for both in a mutually-learning manner. Meanwhile, the extensibility of the above approaches enables our LSM to apply to Faster-RCNN and Deformable-DETR respectively. On the MS-COCO benchmark, our method achieves 3.54\% mAP improvement over state-of-the-art methods under 5\% labeling ratios. 
\end{abstract}
\lfoot{*Corresponding Author}
\section{Introduction}

Deep neural networks~\cite{liu2017survey,kim2020probabilistic} have achieved remarkable progress in the area of object detection. As model complexity increases, a large amount of precisely annotated data is required to train deep networks. To address this need, large-scale object datasets such as MS-COCO~\cite{lin2014microsoft} and Objects365~\cite{shao2019objects365} have been proposed in the community. Nevertheless, the process of annotation can be prohibitively expensive for real-world applications.
\begin{figure}[ht]
    \centering
    \subfigure[\label{fig:pseudo_corr_a}]
	{\includegraphics[width=0.475\linewidth]{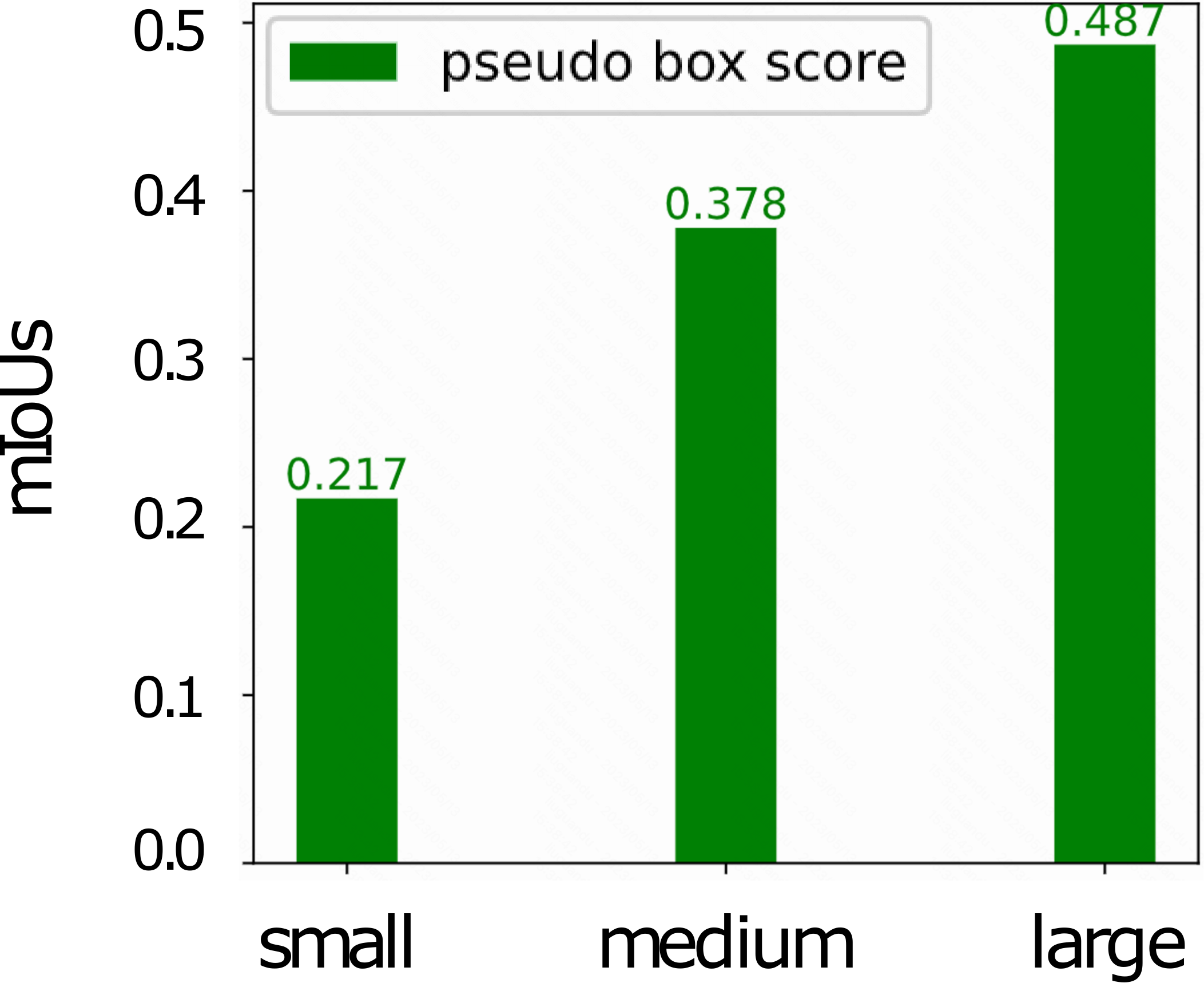}}
	\subfigure[\label{fig:pseudo_corr_b}]
	{\includegraphics[width=0.45\linewidth]{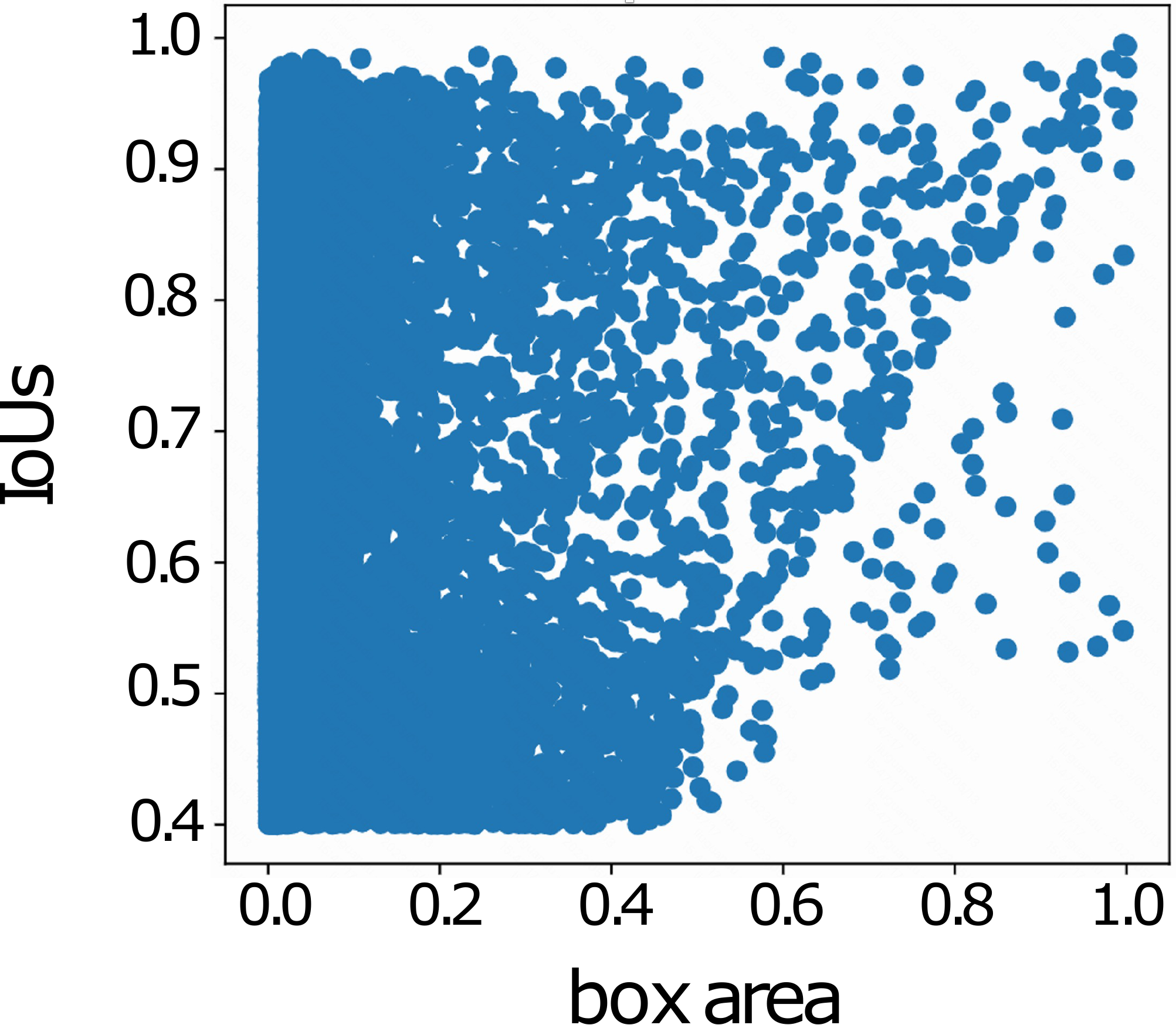}}
    \caption{
        (a) Illustration of the mean IoU for pseudo-labels 
        with three different size intervals (small:[0, 32$\times$32], medium:[32$\times$32, 96$\times$96], large:[96$\times$96, $+\infty$]).
        (b) IoU versus the box area for pseudo-labels.}
    \label{fig:pseudo_corr}
\end{figure}

Recently, semi-supervised object detection (SSOD) has gained attention in the computer vision community~\cite{liu2021unbiased,zhou2021instant,xu2021end}, as it only requires a small amount of annotated data. Most SSOD approaches follow the mean teacher paradigm~\cite{tarvainen2017mean}, which trains a teacher and student model in a mutually beneficial manner. Pioneering approaches such as UBteacher~\cite{liu2021unbiased} and its variants~\cite{zhang2022semi,chen2022label} improve detector performance from the perspective of classificatory balance. To further boost performance, SoftTeacher~\cite{xu2021end} has been proposed, but these approaches do not consider the detailed distribution of pseudo-labels, which contain the box and category.
Recalling previous works, it is apparent that there exists a significant gap in detection performance for boxes of different scales. This observation motivates the following question:
\textit{
Whether there exist differences in the confidence distribution of pseudo-labels between boxes of different scales?
}

\begin{figure}[ht]
    \centering
    \subfigure[\label{pseudo_box_a}]
	{\includegraphics[width=0.39\linewidth]{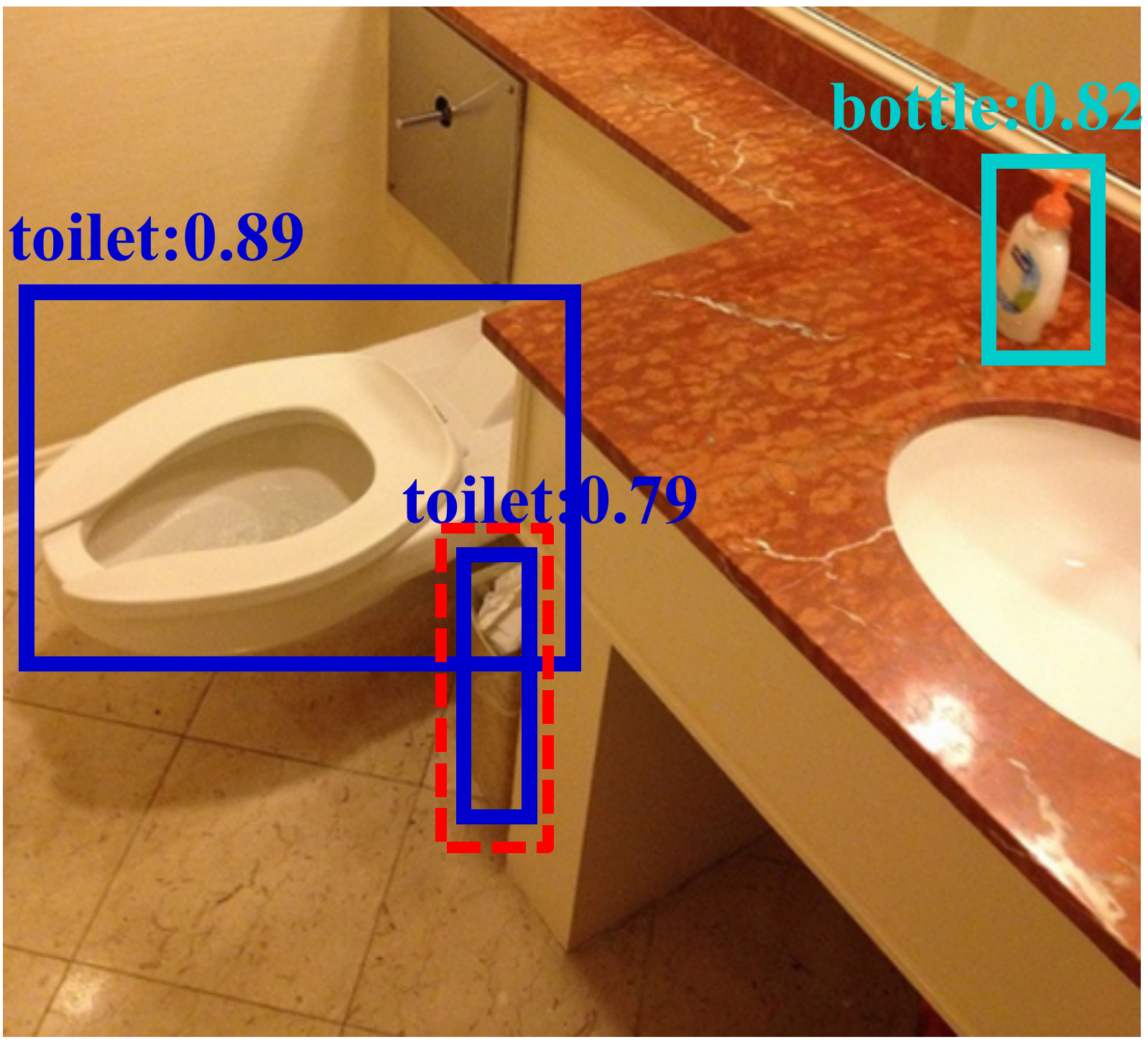}}
	\subfigure[\label{pseudo_box_b}]
	{\includegraphics[width=0.533\linewidth]{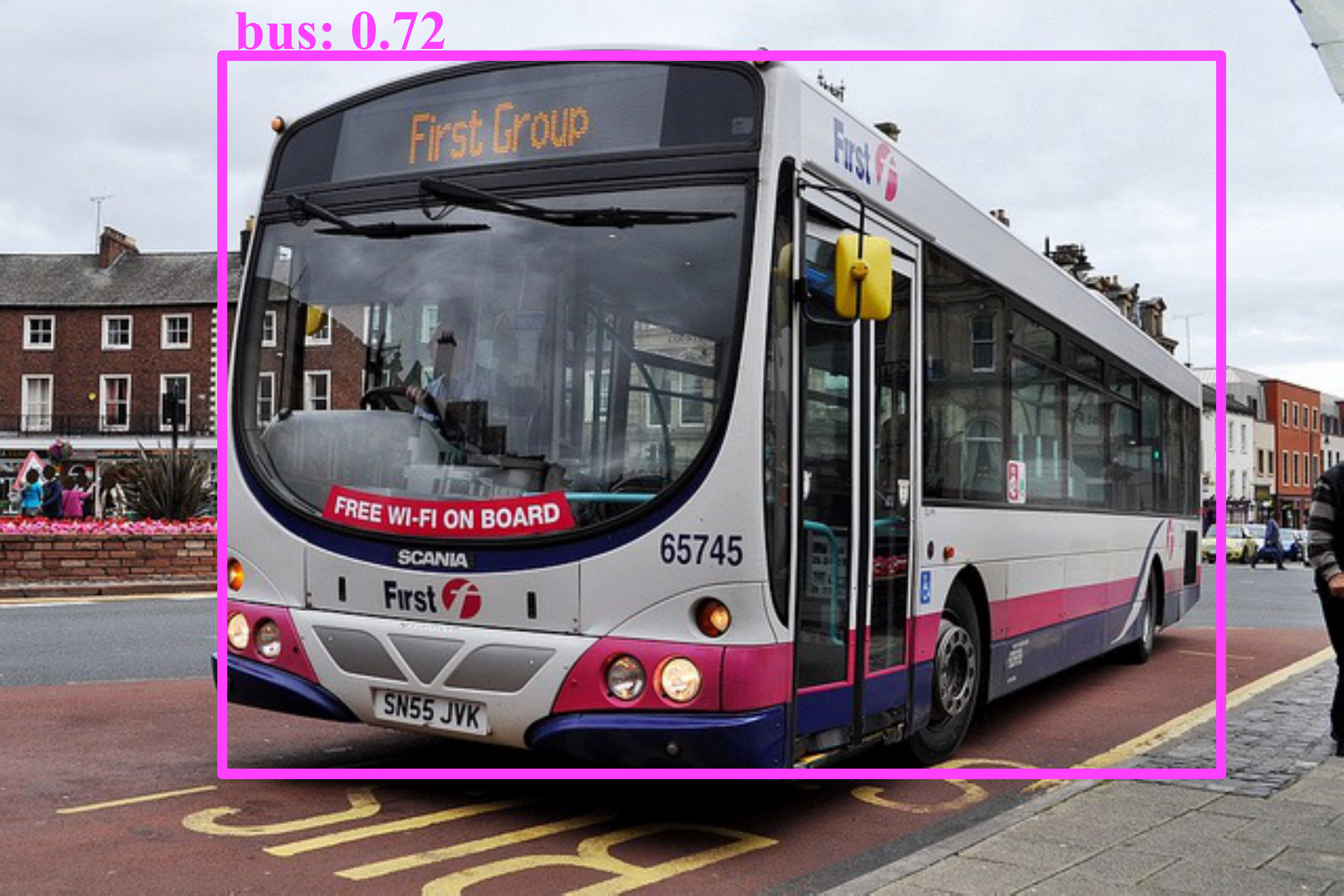}}
    \caption{
        Illustration of pseudo-labels with boundary box, category and classification score. The incorrect pseudo-label (the \texttt{toilet} box in bottom) in (a) has a higher classification score than the correct pseudo-label (\texttt{bus} box) in (b).}
    \label{pseudo_box}
\end{figure}

To address this question, we conducted extensive experiments to explore the relationship between the confidence distribution and the scale of pseudo-labels. Initially, we trained a vanilla Faster-RCNN detector~\cite{ren2015faster} using all the labeled COCO training data. Subsequently, we generated pseudo-labels for the COCO validation set images. Following the methodology of prior works~\cite{he2017mask,cheng2020boundary}, we computed the IoUs (Intersection-over-Union) between the boxes in pseudo-labels and ground-truths with the same category, and we utilized these values to assess the quality of the pseudo-labels. Figure~\ref{fig:pseudo_corr} illustrates the IoU distribution for pseudo-labels of different scales. Figure~\ref{fig:pseudo_corr_a} shows that the pseudo-label quality is positively associated with the box's scale. As the pseudo-label boxes increase in size, there are more pseudo-labels with high quality (Figure~\ref{fig:pseudo_corr_b}), which will provide more precise box and category information for SSOD. Figure~\ref{pseudo_box} visually depicts the pseudo-labels at various scales to qualitatively compare the differences in the confidence distribution of generated pseudo-labels with different scales. We observe that the pseudo-label of \texttt{bus} in Figure~\ref{pseudo_box_b} (i.e., the large pink box with a classification score of $0.72$) is accurate, while the pseudo-label of \texttt{toilet} with a higher score but smaller area (the blue box surrounded by red dotted lines in Figure~\ref{pseudo_box_a}) is incorrect. Intuitively, under the same confidence score, the detector tends to make more accurate predictions for larger-area samples than small-area ones. Hence, leveraging the scale information to exploit low-confidence pseudo-boxes adequately is a valuable technique.

Based on these observations, we have designed a novel training procedure for semi-supervised object detection (SSOD) called Low-confidence Sample Mining (LSM). The direct approach of adding large-area pseudo-labels with low confidence has shown limited improvement (as discussed in Section~\ref{PIM_abla}). Therefore, we propose leveraging low-resolution feature maps to learn reliable large-area candidate boxes, which is more suitable for large-area object training~\cite{singh2018sniper,li2019scale}.
Specifically, LSM introduces an additional branch called pseudo information mining (PIM) for self-learning low-confidence pseudo-labels. PIM downsamples the original image through a feature pyramid network~(FPN) to obtain lower resolution feature maps. A lower threshold is set as DDT~\cite{zheng2022dual} to allow more pseudo-labels to participate in PIM training and help dig hidden credible low-confidence samples. Since scale information PIM uses can be produced in both Faster-RCNN and Deformable-DETR~(DDETR) \cite{zhu2020deformable}, it is natural to introduce DDETR into SSOD.
During the joint training process of the main and PIM branches, we have observed that the candidate boxes learned by both branches have certain complementarity (see Section~\ref{ssec:effect_threshold}). To achieve mutual learning between these two branches, we introduce a self-distillation (SD) module. SD uses the prediction of PIM for low-confidence candidate boxes to supervise the main branch training, and calculates $\mathcal{KL}$ divergence loss between classificatory predictions from the main and PIM branches.

Under the same setting as mean teacher framework~\cite{liu2021unbiased,xu2021end}, our method surpasses the previous state-of-the-arts by significant margins. Especially in the only $5\%$ labeled MS-COCO~\cite{lin2014microsoft}, LSM achieves $3.54\%$ mAP improvement over state-of-the-arts. Furthermore, we find that mean teacher paradigm performances are below baseline on noisy unlabeled data (ImageNet~\cite{deng2009imagenet}). To verify the learning ability of LSM on more-noisy unlabeled data, we conduct a cross-domain task and introduce DDETR baseline into SSOD. Particularly, our method also outperforms DDETR~\cite{zhu2020deformable} by $1.3\%$ mAP in the cross-domain setting.

The contributions of this paper are listed as follows:

\begin{itemize}
\item 
We explore the differences in confidence distribution 
for pseudo-labels between different scales. Moreover, we observe the positive correlation between pseudo-labels area 
and IoUs in SSOD and inspire the use of clean low-confidence boxes from a scale perspective.
These observations provide a new direction to improve SSOD.
\item 
Based on the above observations, 
we propose LSM, 
which uses PIM and SD to exploit clean low-confidence pseudo-labels 
from low-resolution feature maps efficiently.
Extensive experiments are also performed on both Faster-RCNN and DDETR,
which demonstrates that LSM does not rely on specific model components.
\item
We introduce DDETR into SSOD 
and use ImageNet~\cite{deng2009imagenet} as unlabeled data to conduct the cross-domain task, which indicates the excellent denoise capability of LSM.
\end{itemize}

\section{Related Work}
\subsection{Semi-Supervised Learning}
Semi-supervised learning constitutes a fundamental research area within the domain of deep learning. The most prevalent semi-supervised learning approaches are realized through consistency regularization~\cite{izmailov2018averaging,sajjadi2016regularization,kim2022mum,xie2020unsupervised} and pseudo-labeling.
Pseudo-labeling~\cite{sohn2020fixmatch,berthelot2019mixmatch,xie2020self} involves appending predictions to the unlabeled data during the training process, utilizing a teacher model to assist in this task. Consequently, high-confidence predictions are selected as supervisory signals, which serve to enhance model training effectively.
\subsection{Semi-Supervised Object Detection}
The methodologies of semi-supervised object detection (SSOD) primarily stem from semi-supervised learning approaches. STAC~\cite{sohn2020simple} is the first to apply pseudo-labeling and consistency learning to SSOD. It generates pseudo-labels for unlabeled data utilizing a pre-trained model, and subsequently trains a student model with strongly augmented unlabeled images. The mean teacher paradigm~\cite{liu2021unbiased,zhou2021instant,xu2021end} maintains a teacher model for online pseudo-labeling, acquiring reliable pseudo-labels for student model training through a high threshold. However, due to the empirical nature of this threshold, numerous dependable supervisory signals are discarded. Dynamic threshold strategies~\cite{li2022rethinking} seek to obtain a higher quantity of high-confidence supervisory samples by employing a variable threshold.

However, none of the above methods consider the hidden available low-confidence samples. Based on this, recent methods have made efforts in low-confidence samples learning. ~\cite{zheng2022dual} equips vanilla detector framework with the bypass head to learn pseudo-labels with a lower threshold. ~\cite{wangdouble} takes the sum of Top-K probability predictions as the selection basis to expand learning samples. Nevertheless, they all lack further mining that refer to credible information in low-confidence samples and similarly take the incorrect pseudo-labels into training, e.g., the mistaken \texttt{toilet} box in Figure~\ref{pseudo_box_a} will be retained in the above-both methods. Furthermore, the object detector based on transformer has shown powerful performance in recent years. ~\cite{carion2020end,zhu2020deformable} utilize the attention mechanism to get a larger receptive field on the feature map and apply bipartite graph matching to implement end-to-end training. Limited by mean teacher framework, many SSOD methods~\cite{chen2022label,chen2022dense} cannot be directly applied to transformer structure. This hinders the application of Deformable-DETR (DDETR) in SSOD. Owing to the multi-scale feature maps LSM used can be produced by both DDETR and Faster-RCNN. Our work can be applied in DDETR effortlessly.
\subsection{Multi-Scale Invariant Learning}
Multi-scale invariant learning plays a vital role in object detection (OD) by facilitating the learning of objects across different scales. \cite{singh2018sniper} accelerates multi-scale training by sampling low-resolution chips from a multi-scale image pyramid. \cite{li2019scale} employs convolutions with three distinct dilation rates to extract features from objects of varying sizes. Both methods demonstrate remarkable performance in multi-scale learning. Inspired by multi-scale training, our proposed PIM utilizes downsampling and a feature pyramid network (FPN) to generate lower-resolution feature maps for learning large-area objects.

In fact, \cite{li2022pseco} and \cite{guo2022scale} incorporate multi-scale label consistency into the mean teacher framework, striving to learn consistent representations across diverse scales. Although these approaches feature a branch for aligning dense features at different scales, which assists in mining scale-equivariant background features, they still rely on high-confidence pseudo-labels as the training target between the two branches.
While our LSM method will utilize supplementary foreground proposals from low-confidence pseudo-labels.
\begin{figure*}[ht]
    \centering
    \subfigure[\label{overview_a} generation of pseudo-labels]
	{\includegraphics[width=5.6cm]{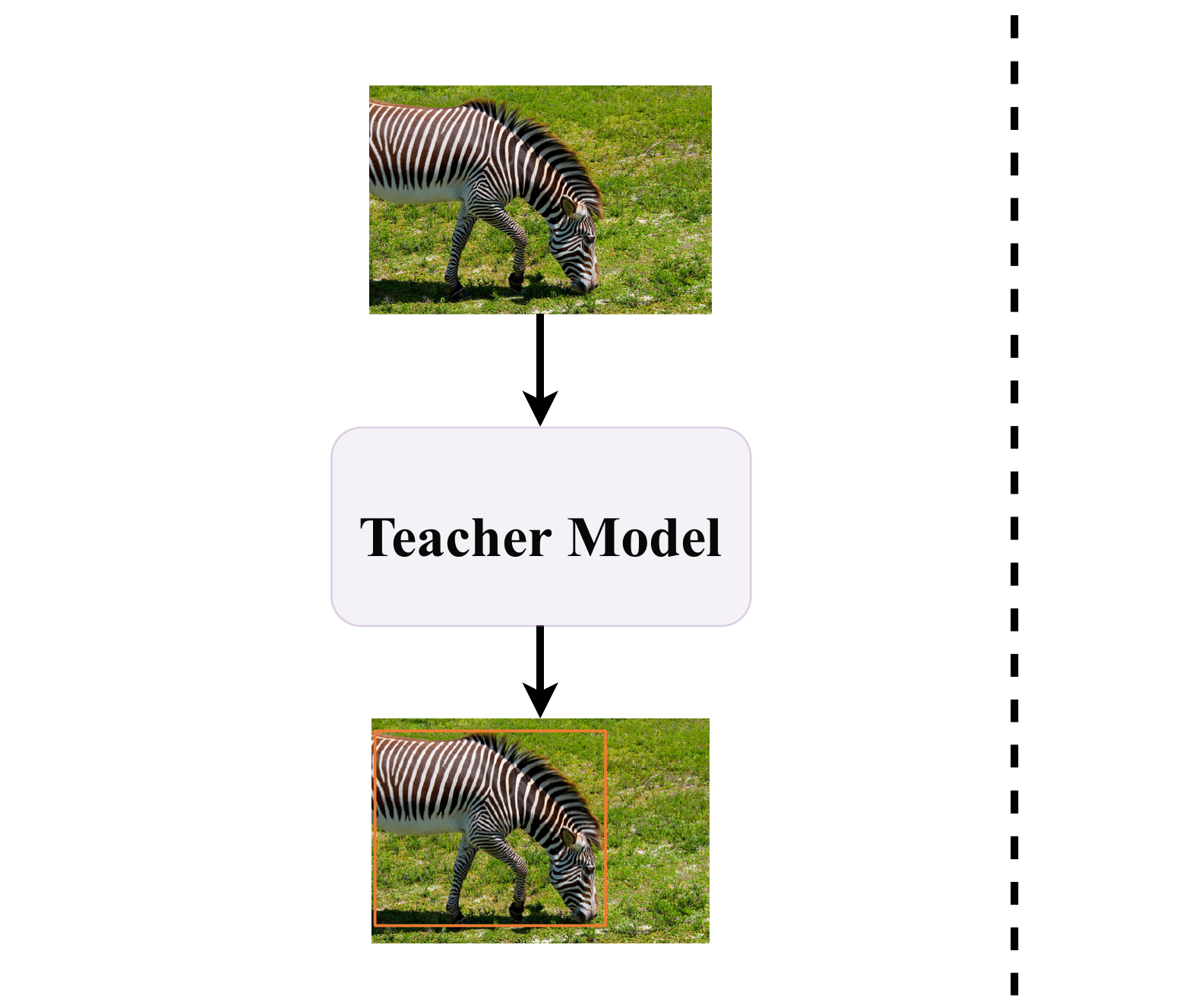}}
	\subfigure[\label{overview_b} low-confidence samples mining (LSM)]
	{\includegraphics[width=12cm]{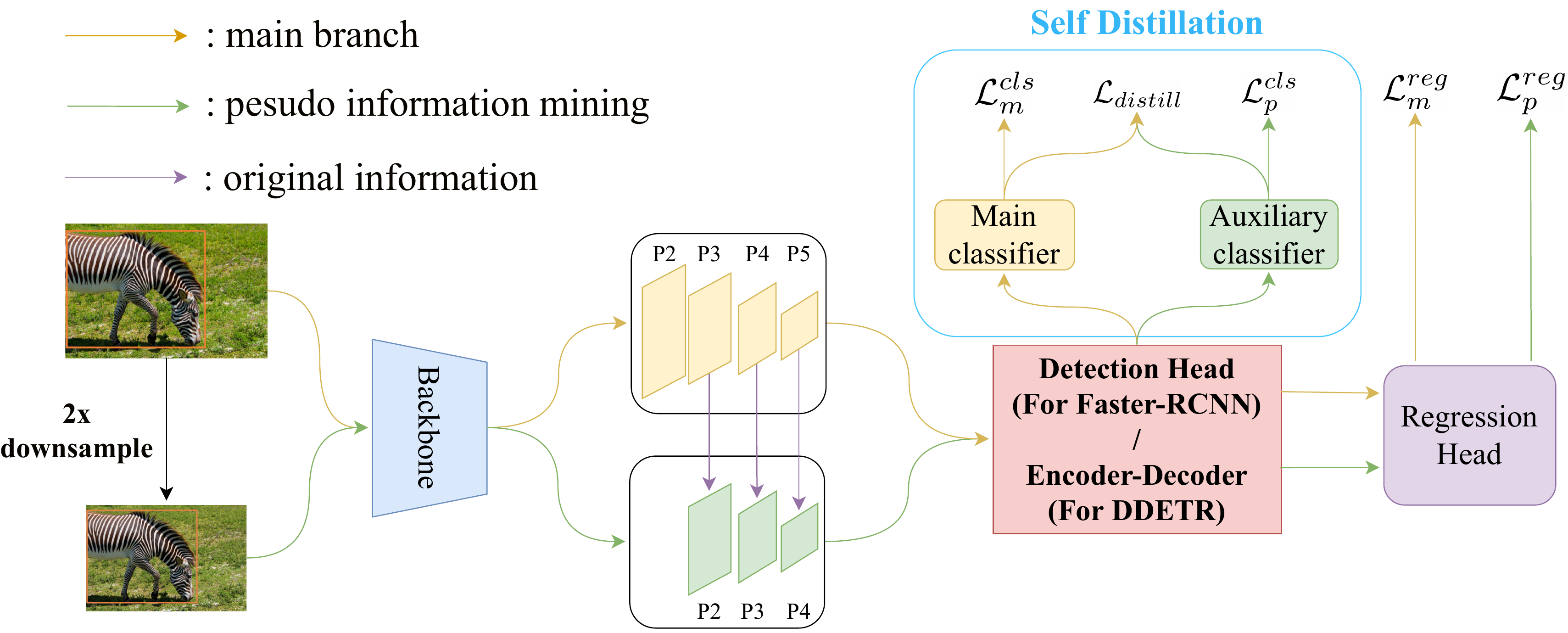}}
    \caption{An overview of our training method. (a) represents the generation of pseudo-labels. (b) illustrates the pipeline of LSM. The yellow line is the main branch following mean teacher framework and the green line is the forward procedure of PIM. The purple line is original information reused by PIM from the main branch. It refers to the proposals generated by the region proposal network (RPN) in Faster-RCNN, and indexes of bipartite graph matching results from in DDETR. The detector is equipped with self-distillation (SD) to learn complementary predictions from two branches.}
    \label{overview}
\end{figure*}

\section{Methodology}
\label{Methodology}
\subsection{Problem Definition}
\label{Section-Problem-Definition}
Semi-supervised object detection aims to use a large amount of unlabeled data to improve model performance, where a small labeled dataset $D^l=\{I_i^l, Y_i^l\}_{i=1}^{N^l}$ and a large unlabeled dataset $D^u=\{I_i^u\}_{i=1}^{N^u}$ are available. $N^l, N^u$ presents the number of labeled, unlabeled data. $Y_i$ contains object information of image $I_i$, including bounding box $\{x_j, y_j, w_j, h_j\}_{j=1}^{N^I_i}$ and category $\{c_j\}_{j=1}^{N^{I}_i}$. ${N^{I}_i}$ presents the number of objects in the $i$th picture.
\subsection{Preliminary: Mean Teacher Framework}
\label{Section-Preliminary}
In the regime of SSOD, this study takes two-stage methods based on the mean teacher paradigm~\cite{tarvainen2017mean} as the baseline. Following previous works, we first train the student model on labeled data and then copy the parameters of the student model to the teacher model. The student model accepts both labeled data and unlabeled data, of which supervision signals come from the teacher model's predictions. The loss function of SSOD can be summarized as supervised loss $\mathcal L_s$ and unsupervised loss $\mathcal L_u$,
\begin{equation}\label{eqn-ssod-loss}
\begin{aligned}
    \mathcal L &= \mathcal L_s + \lambda_u \mathcal L_u,   \\
    \mathcal L_s &= \mathcal L_{cls}(I^s, Y^s) + \mathcal L_{reg}(I^s, Y^s),\\
    \mathcal L_u &= \mathcal L_{cls}(I^u, \hat{Y}^u) + \mathcal L_{reg}(I^u, \hat{Y}^u).\\
\end{aligned}
\end{equation}
{\setlength{\parindent}{0cm}
Among them, $\mathcal L_{cls}$ represents classification loss, and $\mathcal L_{reg}$ represents regression loss. $\hat{Y}^u$ represents the pseudo-labels generated by the teacher model, and $\lambda_u$ is the weight of unsupervised loss. During per iteration, the student model will update the teacher model with its own parameters in the way of exponential moving average (EMA) and generate cleaner pseudo-labels.}
\begin{equation}\label{eqn-ema}
\begin{aligned}
    \theta_s &\gets \theta_s + \frac{\partial \mathcal L}{\partial \theta_s}, \\
    \theta_t &\gets \lambda_{e}\theta_t + (1-\lambda_{e})\theta_s, \\
\end{aligned}    
\end{equation}
{\setlength{\parindent}{0cm}
where $\theta_s, \theta_t$ represents the parameters of the student, teacher model, and $\lambda_e$ represents the ratio of parameter updates.}

\subsection{Low-confidence Samples Mining (LSM)}
\label{Section-LSM}
In this section, we introduce the LSM method. The reliable latent low-confidence pseudo-labels are mined adequately through pseudo information mining (PIM) and self-distillation (SD). The confidence in this section later refers to the classification score.
\subsubsection{Overview}
We introduce the whole pipeline of our LSM method in Figure~\ref{overview} and LSM is applied in the student model after burn-in under the mean teacher framework. First of all, the teacher model trained on a small amount of labeled data $D^l$ is utilized to produce pseudo-labels for $D^u$. Regarding the hybrid training of ground-truths and pseudo-labels, LSM consists of the main branch and pseudo information mining (PIM) for low-confidence pseudo-labels learning. As shown in the yellow line in Figure~\ref{overview}(b), the main branch receives the original feature maps group and leverages pseudo-labels with a high threshold $t$. While the PIM receives downsampling feature maps to learn low-confidence pseudo-labels. The total loss of LSM can be formulated as follow,
\begin{equation}\label{eqn-lsm-loss}
\begin{aligned}
    \mathcal L_{total} = \mathcal L_{m}^{s} + & \lambda_u (\mathcal L_{m}^{u} + \mathcal L_{p}^{u} + \mathcal L_{distill}),\\
\end{aligned}    
\end{equation}
{\setlength{\parindent}{0cm}
where $\mathcal L_{m}^{s}$ and $\mathcal L_{m}^{u}$ refer to supervised loss and unsupervised loss from the main branch, which are the same loss from the mean teacher paradigm. While $\mathcal L_{p}^{u}$ represents unsupervised loss from PIM. $\mathcal L_{distill}$ means distillation loss under the self-distillation (SD) strategy.}
\subsubsection{Pseudo Information Mining (PIM)}
Motivated by the positive correlation between the area and IoUs of pseudo boxes observed in Figure~\ref{fig:pseudo_corr}, we aim to mine reliable pseudo-labels from a scale perspective. As shown in Section~\ref{PIM_abla}, directly incorporating large-area pseudo-labels (area $ > 96\times96$) with a lower-confidence threshold of $0.5$ into training yields some improvement. However, utilizing an area threshold of ($96\times96$) as an empirical parameter is inappropriate for defining large-area boxes. Since detectors tend to learn large objects from low-resolution feature maps, we are inspired to extract valuable large-area pseudo-labels from small-scale images. Consequently, we design PIM from a multi-scale standpoint to learn reliable large-area pseudo-labels with low confidence. The green line in Figure~\ref{overview}(b) depicts the forward procedure of PIM.

Initially, PIM downsamples the original image by a ratio of 0.5 (both width and height are reduced to half of their original sizes), and then inputs the downsampled image into the backbone. A Feature Pyramid Network (FPN) generates a series of small-scale feature maps from the downsampled image, referred to as downsampling feature maps. The $i$th downsampling feature map shares the same size as the ($i+1$)th original feature map. Subsequently, PIM establishes a lower threshold $\alpha$ to include more pseudo-labels in PIM training, fostering the extraction of diverse information by the detector. As downsampling feature maps possess lower resolution, the box features extracted from the detector in PIM are inclined to learn credible large-area candidate boxes. These box features from PIM are then fed into an auxiliary classifier and regression head to compute $\mathcal L_p^{cls}$ and $\mathcal L_p^{reg}$. PIM ensures that the detector learns valuable information from low-confidence pseudo-labels, while the detector's bias towards large-area pseudo-labels on low-resolution feature maps mitigates the adverse impact of noisy small-area pseudo-labels. Moreover, we decrease the loss weight of PIM to further minimize the influence of noisy pseudo-labels.

To reduce the computational load during forward propagation, PIM reuses the original proposals from the main branch (purple line in Figure~\ref{overview}; more details about original information are provided in the supplementary material). Notably, PIM employs $P_2, P_3, P_4$ feature maps to align the $P_3, P_4, P_5$ feature maps from the main branch in a shift alignment manner. As a result, the main branch can share proposals generated by the Region Proposal Network (RPN) with PIM. The loss of PIM can be formulated as follows:
\begin{equation} \label{eqn-Lms}
\begin{aligned}
    \mathcal L_{m}^{s} &= \mathcal L^{cls}_m(F_{main}(E(I_m^s)), Y^s) + \mathcal L^{reg}_m(R(E(I_m^s)), Y^s)
\end{aligned}
\end{equation}
\begin{equation} 
\label{eqn-original-loss}
\begin{aligned}
    \mathcal L_{m}^{u}  &=  \mathbb{I}_{\{\hat{Y}_{score}^u>t\}}[\mathcal L^{cls}_m(F_{main}(E(I_m^u)), \hat{Y}^u) \\
      &+\mathcal L^{reg}_m(R(E(I_m^u)), \hat{Y}^u)]. \\
\end{aligned}
\end{equation}
\begin{equation} \label{eqn-Lpu}
\begin{aligned}
    \mathcal L_{p}^{u} &= \mathbb{I}_{\{\hat{Y}_{score}^u>\alpha\}}[\mathcal L^{cls}_p(F_{aux}(E(I_p^u)), \hat{Y}^u) \\
     &+\mathcal L^{reg}_p(R(E(I^u_p)), \hat{Y}^u)]. \\
\end{aligned}
\end{equation}
Among them, $t$ is the filtering threshold of the main branch. And $\alpha$ is the filtering threshold of PIM, which is lower than $t$. $E$ is the feature extractor, and $R$ is the regression head shared with the main branch and PIM. $F_{main}$ is the main classifier, while $F_{aux}$ is the auxiliary classifier. At the same time, $\mathcal L_p^u$ and $\mathcal L_m^u$ also indicate that the model is forced to learn consistent representations under different-scale features in score interval $[t, \infty]$ to enhance the robustness of the detector model. Eq.~\ref{eqn-Lpu} indicates that PIM can acquire more valuable pseudo boxes from the low-confidence samples due to a lower threshold filtering strategy, especially in Section~\ref{Section-Results} we show that this method can improve recall rate well compared to previous state-of-the-art methods.

PIM has certain similarities with multi-scale label consistency (MLC)~\cite{li2022pseco}. However, MLC aims to improve the robustness of the model via learning the same pseudo-labels from different-scale feature maps, which also ignores low-confidence pseudo boxes. Thereby it interferes the further learning on pseudo-labels. The convincing experimental results are presented in Table~\ref{table:mlc}.
\subsubsection{Self Distillation (SD)}
LSM processes the box predictions from two branches by feeding them into the main classifier and auxiliary classifier, respectively. Due to the complementary predictions observed from the auxiliary classifier, the detector employs self-distillation to incorporate the knowledge of low-confidence bounding boxes learned by the auxiliary classifier into the main classifier. Specifically, we generate categorical predictions using an auxiliary classifier for bounding boxes with confidences in the [$\alpha$, $t$] range. Then, we employ the categorical predictions generated by the main classifier, with confidences in the same interval, to fit the corresponding predictions produced by the auxiliary classifier. The distillation loss is expressed as:
\begin{equation} \label{eqn-L-distill}
    \mathcal L_{distill} = \mathbb{I}_{\{\alpha<\hat{y}^u_{score}<t\}}\mathcal L_{kl}(F_{main}E(I^u_m), F_{aux}E(I^u_p)).
\end{equation}
$\mathcal L_{kl}$ calculates the $\mathcal {KL}$ divergence between the output of the main and auxiliary classifiers. Threshold $t$ and $\alpha$ are the same as those set in Eq.~\ref{eqn-original-loss} and Eq.~\ref{eqn-Lpu}. Regarding the categorical distribution of low-confidence candidate boxes, it is unsuitable to directly choose the category with maximum probability as the hard label due to noise interference. Moreover, some categories with high probability may also be potential labels for the candidate box. Therefore, SD aids the main branch in learning soft predictions from PIM. Additionally, although the auxiliary classifier learns external pseudo-labels, it is still affected by noisy labels. Leveraging the complementarity of dual classifiers, we do not detach the gradient for PIM in SD, allowing the main branch to supervise PIM in a mutually-learning manner. This approach not only ensures that the main classifier learns more pseudo-labels but also mitigates the impact of noisy labels on the auxiliary classifier.

\subsubsection{LSM for Deformable-DETR (DDETR)}

PIM, combined with SD, constitutes the LSM method. Since LSM does not rely on specific network components, it can be effectively transferred to DDETR. The only difference between Faster-RCNN and DDETR, both implemented with LSM, is the original proposals. The results of bipartite graph matching between box predictions from the main branch and pseudo-labels are reused in PIM, as depicted by the purple line in Figure~\ref{overview}(b). Due to memory limitations, we employ the STAC~\cite{sohn2020simple} and pretrain-finetune training strategies for DDETR in the SSOD setting. Specifically, in the first stage, both strategies generate pseudo-labels for unlabeled data using a pre-trained model. In the second stage, STAC trains DDETR with a combination of labeled data and high-confidence unlabeled data, while pretrain-finetune first trains DDETR with unlabeled data and then finetunes it on labeled data. Notably, LSM can be applied in the second stage of both strategies.

In detail, we feed two stacks of original feature maps and low-resolution feature maps into the encoder to obtain two sets of reference points. Then, the object queries conduct cross attention with the reference points from the two sets, respectively, generating two groups of predictions from the two branches. The predictions from the two branches and the pseudo-labels in the two threshold intervals compute $\mathcal L_m^u, \mathcal L_p^u$, respectively. Finally, the dual classifiers generate classification predictions to compute $\mathcal L_{distill}$. During the inference phase, only the main branch is used for forward computation, and the PIM branch is discarded.

\begin{table*}[!ht]  
\centering

\setlength\tabcolsep{8pt}
\resizebox*{1.9\columnwidth}{!}{
\begin{tabular}{c |c|c|c|c|c}
    \toprule
    & \multicolumn{4}{c|}{COCO-standard ($AP_{50:95}$)}  & COCO-additional  \\ 
                        & $1\%$    & $2\%$   & $5\%$ &  $10\%$           & $100\%$ ($AP_{50:95}$)                         \\ \midrule
    Supervised         &  $9.05\pm0.16$ & $12.70\pm0.15$ & $18.47\pm0.22$ & $23.86\pm0.81$ & $40.20$  \\
    CSD~\cite{jeong2019consistency} & $10.51\pm0.06$ & $13.93\pm0.12$ & $18.63\pm0.07$ & $22.46\pm0.08$ & $38.82$  \\
    STAC~\cite{sohn2020simple} & $13.97\pm0.35$ & $18.25\pm0.25$ & $24.38\pm0.12$ & $28.64\pm0.21$ & $39.21$  \\
    Humble Teacher~\cite{tang2021humble} & $16.96\pm0.35$ & $21.74\pm0.24$ & $27.70\pm0.15$ & $31.61\pm0.28$ & $42.17$  \\
    ISMT~\cite{yang2021interactive} & $18.88\pm0.74$ & $22.43\pm0.56$ & $26.37\pm0.24$ & $30.53\pm0.52$ & $39.60$  \\
    Instant Teaching~\cite{zhou2021instant} & $18.05\pm0.15$ & $22.45\pm0.15$ & $26.75\pm0.05$ & $30.40\pm0.05$ & $40.20$  \\ 
    MUM~\cite{kim2022mum} & $21.88\pm0.12$ & $24.84\pm0.10$ & $28.52\pm0.09$ & $31.87\pm0.30$ & $42.11$ \\ 
    \midrule
    UBteacher~\cite{liu2021unbiased} & $20.75\pm0.12$ & $24.30\pm0.07$ & $28.27\pm0.11$ & $31.50\pm0.10$ & $41.30$  \\ 
    UBteacher + LSM        & $\bold{23.95\pm0.02}$ & $\bold{26.60\pm0.04}$ & $\bold{31.97\pm0.09}$ & $\bold{34.75\pm0.13}$ & $\bold{43.23}$  \\ \midrule
    SoftTeacher~\cite{xu2021end} & $20.46\pm0.39$ &       -         & $30.74\pm0.08$ & $34.04\pm0.14$ & $44.50$  \\
    SoftTeacher + LSM     & $\bold{23.76\pm0.18}$ &       -         & $\bold{33.47\pm0.21}$ & $\bold{36.14\pm0.09}$ & $\bold{45.70}$  \\
    \midrule
    PseCo~\cite{li2022pseco} & $22.43\pm0.36$ & $27.77\pm0.18$ & $32.50\pm0.08$ & $36.08\pm0.24$ & $46.10$  \\
    PseCo + LSM & $\bold{24.17\pm0.21}$ & $2\bold{8.96\pm0.07}$ & $\bold{34.21\pm0.11}$ & $\bold{37.33\pm0.08}$ & $\bold{47.01}$ \\
    \hline
\end{tabular}
}
\caption{Comparison with the state-of-the-arts from different percentages of labeled MS-COCO. The margins of error are reported under 5 different random seeds. Where``-'' means the corresponding result is not available.}
\label{table:coco-standard-additional}
\end{table*}

\begin{table}[ht]
\centering
\resizebox{0.8\columnwidth}{!}{
\begin{tabular}{c |c|c}
    \toprule
    & $AP_{50}$ & ${AP_{50:95}}$ \\
    \midrule
    supervised & $72.63$ & $42.13$ \\
    \midrule
    STAC~\cite{sohn2020simple} & $77.45$ & $44.64$ \\
    UBteacher~\cite{liu2021unbiased} & $78.37$ & $50.69$ \\
    Humble Teacher~\cite{tang2021humble} & $80.94$ & $53.04$ \\
    \midrule
    UBteacher + LSM & $\bold{81.61}$ & $\bold{54.90}$ \\
    \hline
\end{tabular}
}
\caption{Comparison with the state-of-the-arts on VOC.}
\label{table:voc}
\end{table}

\begin{table}[ht]  
\centering
\resizebox{1\columnwidth}{!}{
\begin{tabular}{c |c|c}
    \toprule
     & \multicolumn{2}{c}{COCO-ImageNet ($AP_{50:95}$)} \cr
     & Step & mAP \\
    \midrule
    STAC~\cite{sohn2020simple} & $360$K iter & $36.47$ \\
    UBteacher~\cite{liu2021unbiased} & $360$K iter & $38.47$ \\
    UBteacher$^*$~(Ours) & $360$K iter & $\bold{39.87}$ \\
    \midrule
    Deformable-DETR (STAC) & $50$ epoch & $40.11$ \\
    Deformable-DETR$^*$(STAC) & $50$ epoch & $\bold{42.72}$ \\
    \midrule
    Deformable-DETR$^\Omega$~\cite{zhu2020deformable} & $50$ epoch & $43.32$ \\
    Deformable-DETR$^\Phi$ & $50$ epoch & $44.01$ \\
    Deformable-DETR$^\Phi$(LSM) & $50$ epoch & $\bold{45.34}$ \\
    \hline
\end{tabular}
}
\caption{Comparison with the state-of-the-arts on COCO-ImageNet. Where ``*'' represents that the LSM is applied on the corresponding model and ``$\Omega$'' means that the model is only trained on \emph{train2017} in a fully-supervised manner. ``$\Phi$'' indicates that the model is pre-trained with $20\%$ ImageNet, and then finetuned with \emph{train2017}. The last row represents that our method is solely applied in the pre-training stage.} 
\label{table:coco-imagenet}
\end{table}

\begin{table}[ht]  
\small
\centering
\resizebox{1\columnwidth}{!}{
\begin{tabular}{c |c|c|c}
    \toprule
     & Data setting & Step & $AP_{50:95}$ \\
    \midrule
    UBteacher~\cite{liu2021unbiased} & $5\%$ COCO & $180$K iter & $28.27$ \\
    UBteacher$^\Delta$~\cite{li2022pseco} & $5\%$ COCO & $180$K iter & $30.06$ \\
    PIM (In UBteacher) & $5\%$ COCO & $180$K iter & $\bold{31.81}$ \\
    \midrule
    STAC~\cite{sohn2020simple} & COCO-ImageNet & $360$K iter & $36.47$ \\
    STAC$^\Delta$~\cite{li2022pseco} & COCO-ImageNet & $360$K iter & $36.77$ \\
    PIM (In STAC) & COCO-ImageNet & $360$K iter & $\bold{37.87}$ \\
    \midrule
    UBteacher~\cite{liu2021unbiased} & COCO-ImageNet & $360$K iter & $38.47$ \\
    UBteacher$^\Delta$~\cite{li2022pseco} & COCO-ImageNet & $360$K iter & $38.81$ \\
    PIM (In UBteacher) & COCO-ImageNet & $360$K iter & $\bold{39.87}$ \\
    \hline
\end{tabular}
}
\caption{Comparison with Multi-scale Label Consistency (MLC). ``$\Delta$'' indicates that the model is trained using MLC.
}
\label{table:mlc}
\end{table}
\section{Experiment}
\subsection{Datasets} 


In this section, we carry out extensive experiments to validate the effectiveness of LSM on the MS-COCO~\cite{lin2014microsoft}, PASCAL VOC~\cite{everingham2010pascal}, and ImageNet~\cite{deng2009imagenet} benchmarks.

MS-COCO contains two training sets, the \emph{train2017} dataset with 118K labeled images and the \emph{unlabeled2017} dataset with 123K unlabeled images. Following previous methods, we conduct experiments under three settings: (1) \textbf{COCO-standard}: we sample 1\%, 2\%, 5\%, and 10\% of the images from \emph{train2017} as labeled data, while the rest are treated as unlabeled data. (2) \textbf{COCO-additional}: We use the full \emph{train2017} dataset as labeled data and the \emph{unlabeled2017} dataset as unlabeled data. (3) \textbf{VOC}: We use the \emph{VOC07-trainval} as the labeled dataset and the \emph{VOC12-trainval} as the unlabeled dataset. We evaluate the model on \emph{COCO-val2017} for (1)(2) and \emph{VOC07-test} for (3).

In addition to these three traditional settings, we find that previous SSOD methods have not conducted cross-domain experiments on a more noisy unlabeled dataset. To demonstrate the denoising capacity of LSM-equipped Deformable-DETR on cross-domain tasks under SSOD settings, we introduce a fourth experimental setting: (4) \textbf{COCO-ImageNet}: We use the full \emph{train2017} dataset as labeled data and randomly choose 20\% of ImageNet as noisy unlabeled data. The pseudo-labels for unlabeled data are predicted by the Faster-RCNN trained on the \emph{train2017} dataset.

\subsection{Implementation Details}
To be fair, we use Faster-RCNN as our base object detector as same as previous studies~\cite{liu2021unbiased,xu2021end}. The weights of the backbone are initialized with ImageNet pre-trained model. For the main branch, we set pseudo boxes filtering threshold $t$ to $0.7$. While for LSM, which can have a higher tolerance for pseudo boxes, we set the threshold $\alpha$ to $0.5$. In all training settings, each of our training batches follows previous correspond works. For COCO-standard, the entire training steps are $180,000$, of which the first $20,000$ steps are used to pre-train the student model with labeled images. For COCO-additional, pre-training steps are $90,000$, and the whole training steps are $360,000$. For COCO-ImageNet, it takes the same training steps as COCO-additional due to the data size 
is close. In our experiments, strong data augmentation involves random jittering, gaussian noise, crop, and weak data augmentation involves random resize and flip. Moreover, we follow the existing work~\cite{liu2021unbiased,xu2021end} to set the above hyperparameters.

\begin{figure}[ht]
    \centering
    \subfigure[\label{recall_a} 5\% COCO Avg Recall]
	{\includegraphics[width=0.46\linewidth]{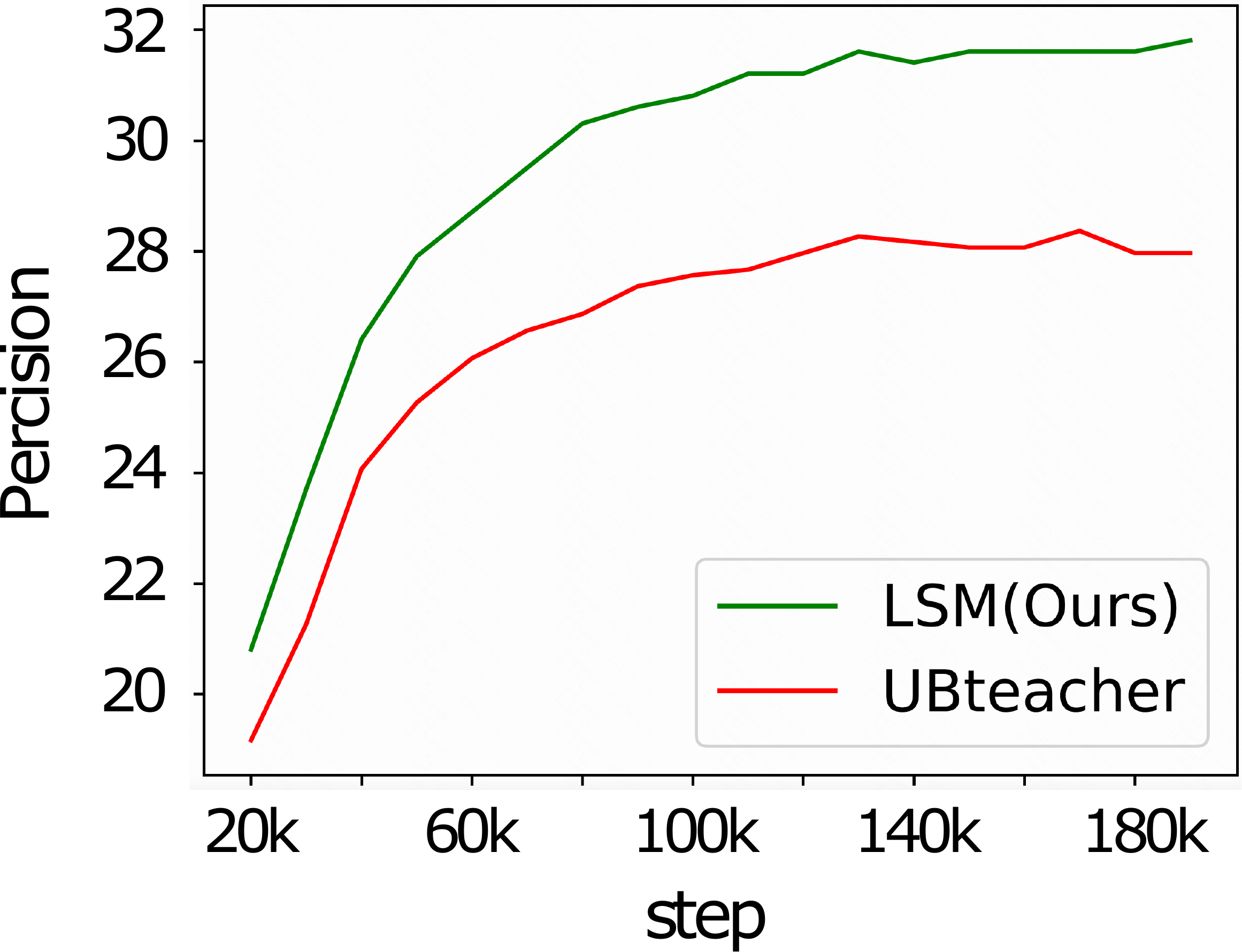}}
	\subfigure[\label{recall_b} 5\% COCO Avg Precision]
	{\includegraphics[width=0.46\linewidth]{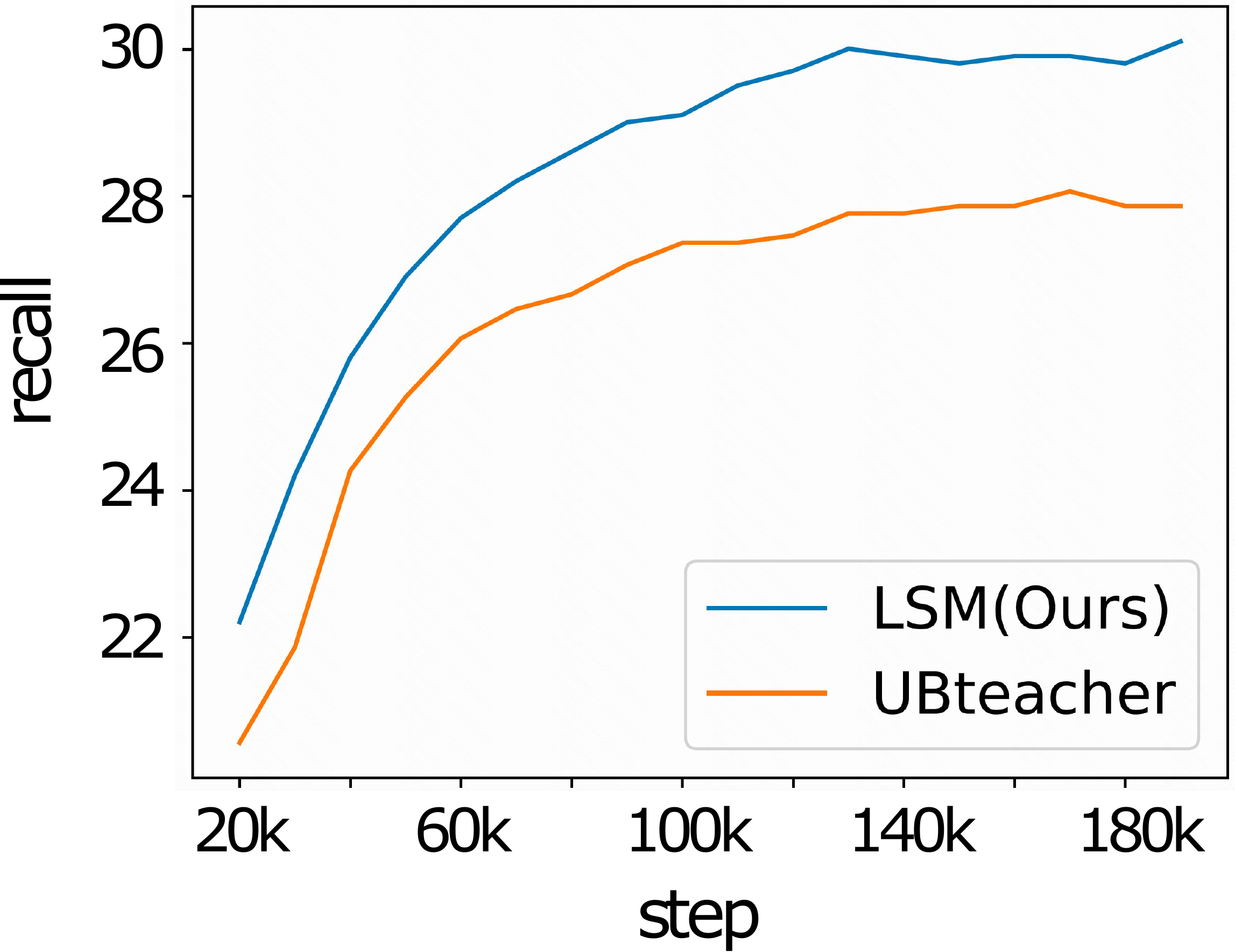}}
    \caption{
        Comparison of the average recall and average precision curves for UBTeacher and LSM(ours) on 5\% COCO-standard.}
    \label{recall}
\end{figure}
\subsection{Results}\label{Section-Results}
\subsubsection{COCO-standard}
We first evaluate our method under the COCO-standard setting. As shown in Table~\ref{table:coco-standard-additional}, UBteacher~\cite{liu2021unbiased} equipped with LSM can perform better than previous work. When trained on $5\%$ COCO-standard, LSM outperforms the UBteacher by $3.54\%$ mAP. Even if LSM is applied to SoftTeacher (PseCo), it can also improve by $2.71\%$ ($1.47\%$) mAP on average in $1\%$, $5\%$ and $10\%$ labeled data. We attribute the success of model performance to the stronger ability to capture object boxes in LSM. As shown in Figure~\ref{recall_a}, under the setting of $5\%$ COCO-standard, the recall rate of LSM is higher than that of the UBteacher in the whole training stage, which means that after adding LSM, the model can better detect the previous missing boxes. Figure~\ref{recall_b} indicates that the mAP metric of LSM is better than that of UBTeacher. Theoretically, LSM adds an extra branch to learn low-confidence pseudo-labels, and the downsampling operation biases the model to learn more clean large-area pseudo boxes thus extracting additional information.
\subsubsection{COCO-additional}
In this section, we verify LSM can be further improved when trained on large-scale labeled data with additional unlabeled data. As shown in Table~\ref{table:coco-standard-additional}, when LSM is applied to the UBTeacher, it can improve $1.93\%$ mAP compared to the UBTeacher baseline 
While LSM also achieves $1.2\%$ mAP improvement with applying in SoftTeacher baseline. These results indicate that our method achieves satisfactory improvement on large-scale unlabeled datasets.
\subsubsection{VOC}
We evaluate models on a balanced dataset \textbf{VOC} to demostrate the generalization of LSM. Table~\ref{table:voc} provides the mAP results of CSD, STAC, UBteacher, Humble Teacher and our LSM-equipped UBteacher. Our method achieves $4.21\%$ mAP improvement compared with UBteacher baseline and $1.86\%$ mAP improvement compared with Humble teacher. Our method surpasses the other state-of-the-art results with a large margin. These results demonstrate that LSM can improve the existing SSOD consistently in various datasets.
\subsubsection{COCO-ImageNet}
To verify the effectiveness of LSM-equipped DDETR, we propose a new cross-domain setting: COCO-ImageNet. Considering that DDETR converges slowly, we use epochs as the unit in the training process.

As shown in Table~\ref{table:coco-imagenet} row 2, the detector reduces $1.73\%$ mAP compared to the fully supervised mode in UBteacher paradigm.
While LSM demonstrates excellent denoise capability, with an improvement of $1.4\%$ mAP compared to the UBTeacher baseline. 
Furthermore, in the training mode of pretrain-finetune, we find that DDETR performs better ($0.7\%$ mAP) than the supervised baseline, which indicates that the pretrain-finetune mode can better utilize more noisy pseudo-labels.
Moreover, after applying LSM to the pre-training stage of DDETR, we observe that the model can achieve a $1.3\%$ mAP improvement. This shows that LSM not only can be applied to Faster-RCNN and DDETR as a decoupling method but also has excellent learning ability in noisy labels.
\subsubsection{Compared with Multi-scale Label Consistency (MLC)}
The downsampling method used by our PIM follows the multi-scale label consistency method. MLC is widely used in object detection as an incremental method.
 However, existing methods force the model to learn a consistent representation of high-confidence pseido-labels between the two branches. PIM, on the other hand, equips the downsampling branch with a lower filtering threshold, to capture more information from pseudo-labels. To verify that our PIM outperforms the MLC method, we apply these two methods under two settings of $5\%$ COCO-standard and COCO-ImageNet, respectively.

As shown in Table~\ref{table:mlc}, under the setting of $5\%$ COCO, applying the MLC on UBteacher can improve $1.79\%$ mAP, while applying PIM can improve $3.54\%$ mAP. In the COCO-ImageNet setting, we find that applying MLC on STAC brings a limited improvement ($0.3\%$ mAP) while applying the PIM can bring $1.4\%$ mAP improvement. 

\section{Ablation Study}
\subsection{Effects of Pseudo Information Mining Branch}
\label{PIM_abla}
PIM uses downsampling method to obtain three different-resolution feature maps of $P_2^d$, $P_3^d$, and $P_4^d$ generated by feature pyramid network (FPN). As shown in Table~\ref{table:ablation-coco-standard}, we select multiple combinations from three feature maps to learn low-confidence samples. From row 2, the baseline has a certain improvement ($0.76\%$ mAP) through directly adding large-area pseudo-labels exceeding a lower threshold to the training. Whereas we find that using $P_2^d$, $P_3^d$, and $P_4^d$ simultaneously in the PIM, the model performs the best, $3.2\%$ mAP higher than the UBteacher baseline in row 6. As shown in row 3, if we only use the $P_2^d$, $P_3^d$, we find that the extra object information learned by the PIM is very limited, which is only $0.4\%$ mAP higher than the baseline. When we add the lower resolution feature map $P_4^d$ (as shown in row 4), we find that the performance will be significantly improved, which is $2.41\%$ mAP higher than the baseline. Through the comparison of the row 3 and the row 5 of Table~\ref{table:ablation-coco-standard}, we can find that the combined detection of $P_3^d$ and $P_4^d$ on large objects is $3.54\%$ $AP_L$ higher than that of $P_2^d$ and $P_3^d$. This shows that using lower resolution feature maps for PIM can indeed better mine large objects with lower confidence. 
\begin{table}[h!]  
\small
\centering
\resizebox{1\columnwidth}{!}{
\begin{tabular}{c|ccc |c|ccc}
    \toprule
    & $P_2^d$ & $P_3^d$ & $P_4^d$ & $AP_{50:95}$ & $AP_S$ & $AP_M$ & $AP_L$ \\
    \midrule
    $1$ &  &  &   & $20.75$ & $9.21$ & $21.73$ & $27.32$ \\
    $2$ &  &  &   & $21.41$ & $9.23$ & $22.45$ & $28.13$ \\
    \midrule
    $3$ & \Checkmark & \Checkmark &   & $21.15{(+0.4)}$ & $9.10$ & $22.63$ & $27.87$ \\
    $4$ & \Checkmark & & \Checkmark & $23.16{(+2.41)}$ & $11.36$ & $25.73$ & $30.91$ \\
    $5$ & & \Checkmark & \Checkmark & $23.24{(+2.49)}$ & $11.48$ & $25.88$ & $\bold{31.41}$\\
    $6$ & \Checkmark & \Checkmark& \Checkmark & $\bold{23.95}{(+3.2)}$ & $\bold{11.90}$ & $\bold{26.83}$ & $31.21$ \\
    \hline
\end{tabular}
}
\caption{Ablation study on PIM under $1\%$ COCO-standard. The row 1 represents the UBteacher baseline without using $P_2^d, P_3^d, P_4^d$. The row 2 indicates that large objects exceeding a lower threshold ($t>0.5$, area $>96\times96$) are added to the training of UBteacher baseline.}
\label{table:ablation-coco-standard}
\end{table}
\subsection{Effects of Filter Threshold}
\begin{figure}
\centering
\includegraphics[width=0.38\textwidth]{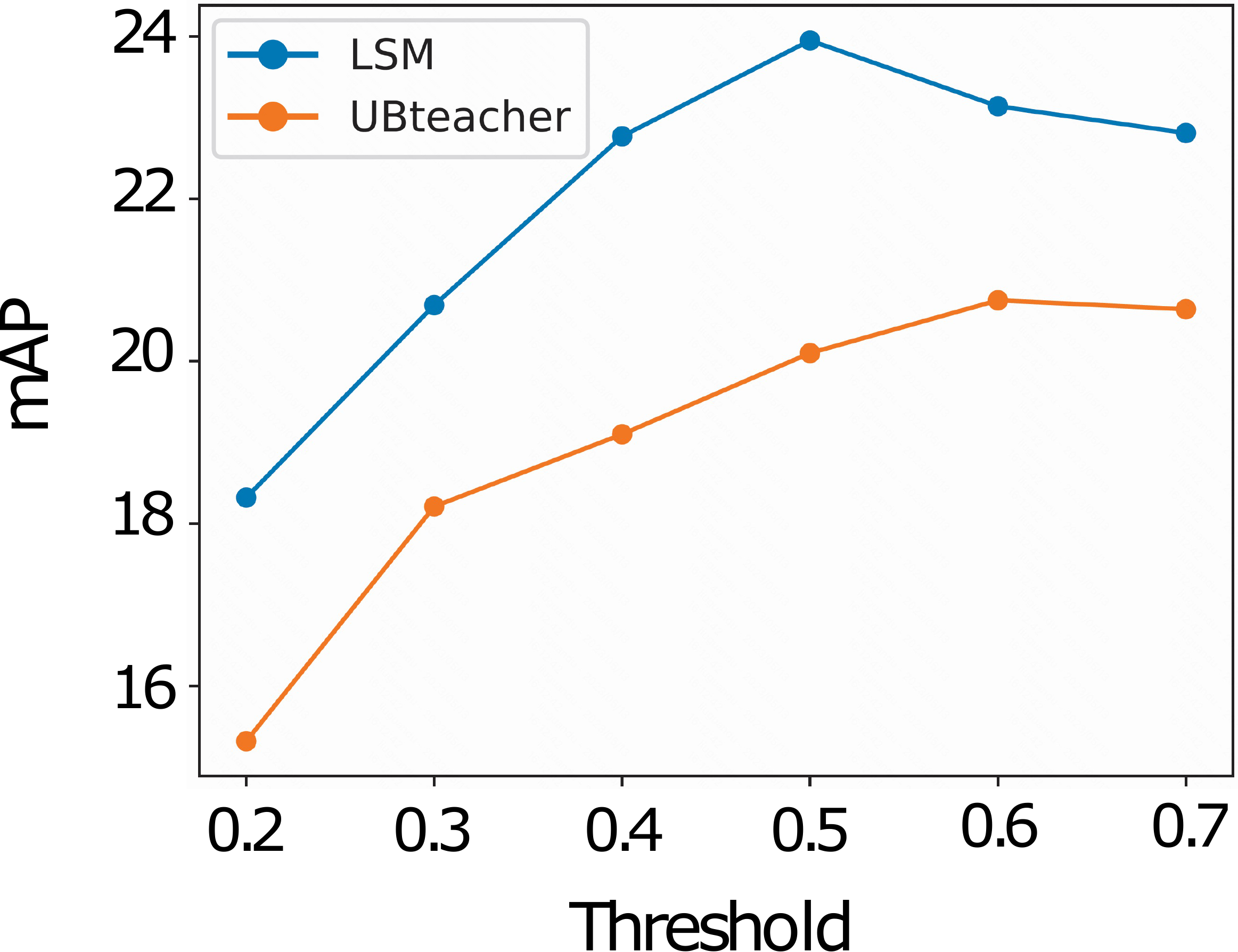}
\centering\caption{\label{threshold}Ablation study on filtering threshold}
\end{figure}

Threshold plays a key role in screening high-quality pseudo-labels. Figure~\ref{threshold} shows the performance of the model under $1\%$ COCO-standard at different thresholds. The red line represents the corresponding performance of the UBteacher after adjusting the threshold $t$. The blue line shows the corresponding performance of the LSM-equipped UBteacher after adjusting the threshold $\alpha$. For UBTeacher, it is difficult for the model to utilize the useful low-confidence pseudo-labels, and the performance of the model becomes worse as the threshold decreases. Moreover, it is difficult for the model to improve further after the threshold exceeds $0.6$. For LSM, the performance of the model reaches the highest $23.95\%$ mAP for $\alpha=0.5$. Therefore, our method can make sufficient use of pseudo-labels with low confidence (i.e., $\hat{Y}_{score}^u \in [0.5, 0.7])$, which is not achieved by previous methods.
\subsection{Effects of Self-distillation}
\label{ssec:effect_threshold}

In this experiment, 
the original PIM and the PIM equipped with SD are compared.
Table~\ref{table:ablation-sd} 
shows the performance gain on the UBTeacher ($1\%$ COCO-standard setting) and DDETR models respectively.
As can be seen,
PIM+SD can improve $0.2\%$ (resp. $0.4\%$) mAP on UBTeacher (resp. DDETR) than PIM,
which demonstrates the effectiveness of SD.
As the PIM branch has learned external low-confidence pseudo-labels,
the PIM branch would be complementary to the main branch. This complementarity is illustrated with more visual results in the supplementary material
\begin{table}[h!]  
\small
\centering
\resizebox{0.8\columnwidth}{!}{
\begin{tabular}{c|cc|cc}
\toprule
    & \multicolumn{2}{c|}{UB} & \multicolumn{2}{c}{DDETR} \\
    \midrule
    Method & PIM & PIM+SD & PIM & PIM+SD \\
    \midrule
    $AP_{50:95}$ &  $34.55$ & $\bold{34.75}$ & $44.9$ & $\bold{45.3}$\\
    \hline
\end{tabular}
}
\caption{Ablation study on self-distillation.} 
\label{table:ablation-sd}
\end{table}
\section{Conclusion}
In this study, we dive into the problem of discarding numerous low-confidence samples. Motivated by the positive correlation between area and IoUs of pseudo boxes, we propose the LSM method consists of PIM and SD. As high-level feature maps is conductive to learn large candidate boxes, PIM utilizes downsampling method and a lower threshold to extract diverse information from low-confidence pseudo-labels. Moreover, LSM takes advantage of SD to make PIM and main branch in mutually-learning manner. Sufficient experiments on benchmark demonstrate the superiority of our method. At the same time, our method can be freely applied to DETR framework, and shows excellent denoise ability on the cross-domain task.

\section*{Acknowledgments}
This work was supported by the NSFC under Grant 62072271. Jun-Hai Yong was supported by the NSFC under Grant 62021002.

\bibliographystyle{named}
\bibliography{ijcai23}

\end{document}